# An Attention-based Method for Action Unit Detection at the 3rd ABAW Competition

Duy Le Hoai, Eunchae Lim, Eunbin Choi, Sieun Kim, Sudarshan Pant, Guee-Sang Lee, Soo-Huyng Kim, Hyung-Jeong Yang[1]

Department of Artificial Intelligence Convergence, Chonnam National University
Gwangju 61186, South Korea

hoaiduy1396@gmail.com, enechae78@gmail.com, iidmsqlss@gmail.com, leaza34@gmail.com,
sudarshan@chonnam.ac.kr, {gslee, shkim, hjyang}@jnu.ac.kr

## Abstract

*Facial Action Coding System is an approach for modeling the complexity of human emotional expression. Automatic action unit (AU) detection is a crucial research area in human-computer interaction. This paper describes our submission to the third Affective Behavior Analysis in-the-wild (ABAW) competition 2022. We proposed a method for detecting facial action units in the video. At the first stage, a lightweight CNN-based feature extractor is employed to extract the feature map from each video frame. Then, an attention module is applied to refine the attention map. The attention encoded vector is derived using a weighted sum of the feature map and the attention scores later. Finally, the sigmoid function is used at the output layer to make the prediction suitable for multi-label AUs detection. We achieved a macro F1 score of 0.48 on the ABAW challenge validation set compared to 0.39 from the baseline model.*

1. **Introduction**

Facial expression recognition (FER) is an important task in artificial emotional intelligence (AEI) research as it not only helps to identify human affective states but also allows to mimic various emotions during human-machine communication. Facial expression is one of the common ways to express emotions in humans. Ekman developed the Facial Action Coding System (FACS) [11] including 46 action units that match expressions and human emotions. Action Units (AU), the points related to specific facial muscle actions are related to the facial expressions [12].

AU detection has become an important facial analysis task that has been applied to various areas such as the healthcare and entertainment industry by extracting and recognizing features representing human emotions [3].

However, due to the scarcity of large datasets recognition of facial expressions is still a challenge [12]. The third Affective Behavior Analysis in-the-wild (ABAW) 2022 competition [12-21] provides a benchmark and a massive dataset based on the large scale in-the-wild Aff-Wild2 database for four challenges including Valence-Arousal Estimation, Expression Classification, Action Unit Detection, and Multi-Task Learning.

In this paper, we introduce our approach for Action Unit Detection challenge in the ABAW 2022 competition. First, we enforce a lightweight feature extractor to extract visual information from image frames. Second, we leverage the attention mechanism to capture the important intra-region in the face image to improve the AU detection performance. In addition, to counter the imbalance data problem, we execute class reweights binary cross-entropy loss. In the next session, we will explain our method in detail.

## 2. Related Works

### 2.1. Facial Emotion Recognition

FER methods can be classified into hand-crafted feature-based, temporal feature-based, and deep learning-based methods according to the extracted features. Hand-crafted features include texture-based features such as Gabor filter [22], SIFT [23], and HOG [24], geometry-based features, and hybrid features. Temporal feature-based methods use temporal features from the face appearance related to AUs. Li et al. [25] used the Long Short-Term Memory (LSTM) network to detect AUs. Deep learning-based methods have been studied for large-scale data sets and better accuracy of AU detection. Wang et al. [26] proposed Self-Cure Network (SCN), a deep network that avoids over-fitting from uncertain facial images.

---

[1] Corresponding Author.

## 2.2. Action Unit Detection Analysis

In the ABAW2 competition so far, many teams proposed single-task and multi-task learning methods to recognize AUs. In this section, we briefly introduce a single task learning method for recognizing AUs. Pahl *et al.* [27] propose a method based on multi-label class balancing algorithm for unbalanced datasets and a ResNet with 18 layers. Ji *et al.* [28] use an end-to-end deep neural network to carry out the static and dynamic features extraction and the AU classification. Zhang *et al.* [29] present a method based on JAA-Net and Graph Convolutional Network. Saito *et al.* [30] introduce a method of inserting siamese-based networks, known as uncertainty models, into automatic AUs recognition methods using a pairwise deep architecture [31] to further reduce AUs recognition errors.

## 3. Proposed Method

The overall of our proposal is illustrated in Figure 1. Our approach comprises two main components: a feature extractor for extracting visual features from the input images and an attention module to induct the model focusing on the important local regions for predicting action units. At the output layer, we leverage the sigmoid function to adapt with the multi-label classification task of the action unit detection challenge.

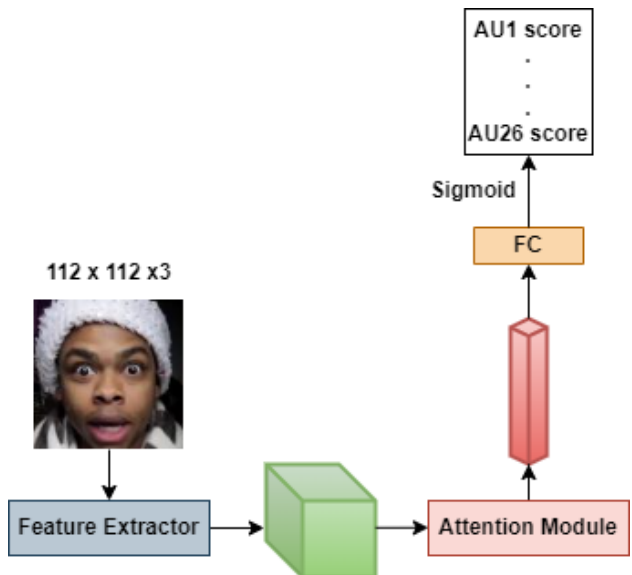

Figure 1. An overview of our proposed model.

### 3.1 Model Architecture

**Feature Extractor** Previous approaches [4,5,6] make use of the pre-trained state-of-the-art models on the large-scale datasets as a feature extractor. In this work, we used a custom lightweight feature extractor. We simply construct a convolutional block that contains a plain convolutional (Conv) layer followed by a Batch Normalization layer with ReLU as activation function and a max-pooling layer as the last layer of the block. We utilize a 3 x 3 kernel with a stride is 1 for all convolutional layers and strides of 2 x 2 for all max-pooling layers. The face image of size 112 x 112 x 3 is passed through a stack of 6 convolutional blocks with the number of filters for each block being 32, 64, 128, 128, 256, 256 sequentially. The extracted visual feature is a 256-dimensional feature map.

**Attention Module** It is well known that the attention mechanism has been established a significant impact in many research fields such as natural processing [7], computer vision [8], speech recognition [9], etc. We adopt the same mechanism as described in [10], Figure 2, to guide the network selectively focus on local salient parts to capture meaningful visual features for action unit detection task. First, the feature map is split into a set of W*H sub-vectors, each vector $v_{i,j}$ has C elements corresponding to each location in the feature map. Next, a fully connected (FC) layer followed by Batch Normalization and ReLU layer is added. Then a 1-unit FC layer is used to compute the score value $s_{i,j}$ for each sub-vector. After that, the weight $w_{i,j}$ of each sub-vector is calculated by a Softmax function. Finally, the attention encoded vector is obtained as a weighted sum of these sub-vector:

$$attention\ vector = \sum_{i=0}^{H}\sum_{j=0}^{W} w_{i,j} v_{i,j} \quad (1)$$

A feed-forward layer with a sigmoid activation function is adopted to produce the final action unit prediction.

### 3.2 Loss Function

Action Unit Detection challenge is a multi-label classification problem. Binary Cross-Entropy (BCE) is usually used for multi-label classification tasks with a sigmoid function as an activation function in the output layer. To solve the imbalance class problem in the dataset, we introduce the sample weights in the BCE as follow:

$$\mathcal{L}(y, \bar{y}) = \frac{1}{C}\sum_{i=1}^{C} \mathcal{L}_{BCE}(y_i, \bar{y}_i) \quad (2)$$

$$\mathcal{L}_{BCE}(y_i, \bar{y}_i) = -[w_i y_i . \log \bar{y}_i + (1 - y_i).\log(1 - \bar{y}_i)] \quad (3)$$

$$w_i = \frac{\#\ total\ training\ samples}{2 * (\#\ positive\ sample\ in\ i-th\ AU)} \quad (4)$$

Where, $y$ and $\bar{y}$ denote the ground truth and prediction, respectively. Variable C denotes the number of action units. In this work, C is equal to 12.

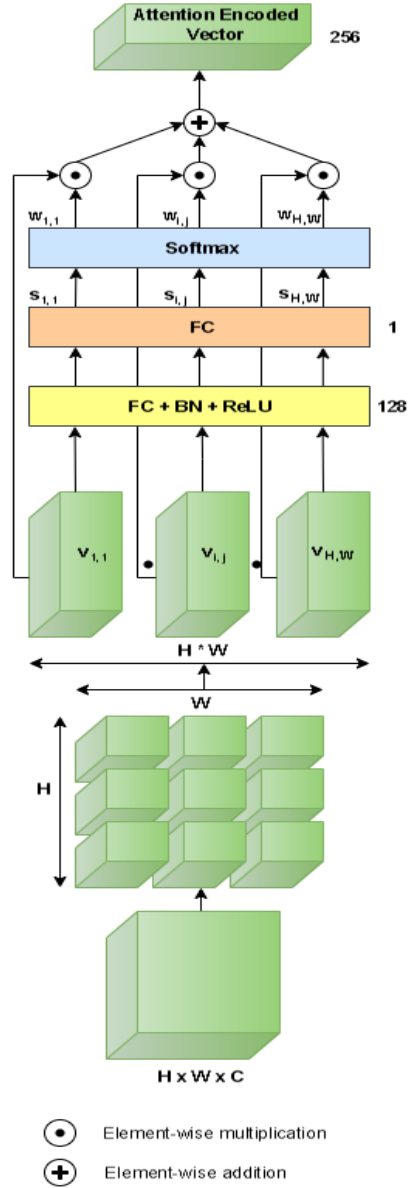

Figure 2. The details of the attention module that is based on the work [10].

## 4. Experiments

### 4.1 Dataset and Metric

The proposed method was trained and validated on 547 videos that contain annotations in terms of 12 AUs. We directly used the face-cropped and aligned images provided by the ABAW organizer. These images have the size of 122x122x3 in RGB color space. The performance metric in the competition is the macro F1 score across all 12 AUs:

$$P_{AU} = \frac{\sum_{au} F_1^{au}}{12} \quad (5)$$

### 4.2 Experiment Setting

Our model is implemented using TensorFlow [1] with an NVIDIA RTX 3080 graphics card. We used Adam [2] optimizer with the initial learning rate of 0.001 which is set to 0.0001 after 5 epochs, the hyper-parameters $\beta_1 = 0.9$ and $\beta_2 = 0.999$. The mini-batch size is set to 256.

### 4.3 Results

Our experiment result on the validation set is shown in Table 1. We achieved an F1 score of 0.48 compared to 0.39 of baseline result on the validation set. It indicates that our approach outperforms the baseline method.

| Method | Macro F1 score |
|---|---|
| Baseline [13] | 0.39 |
| **Ours** | **0.48** |

Table 1. AU detection result on the validation set. Our best result is indicated in bold.

## 5. Conclusion

In this work, we present a deep learning-based approach using attention mechanism and class re-weight loss for the Action Unit Detection challenge of the ABAW2022 competition. The experiment result shows that the attention mechanism is an effective technique for AU detection task and class re-weight can tackle the imbalance dataset problem in the multi-label classification task. Our method outperforms the baseline model on the validation set with an F1 score as the model performance metric.